\documentclass{article}
\usepackage{hyperref}
\usepackage{spconf,amsmath,graphicx}
\usepackage{float}
\usepackage{balance}
\usepackage{cite}
\usepackage{caption}
\usepackage{subfigure,multirow}
\usepackage{booktabs}
\usepackage{array}
\usepackage{framed}
\usepackage{bm}

\usepackage{setspace}
\usepackage{epstopdf}
\usepackage{color}
\usepackage{xcolor}

\title{PFT-SSR: Parallax Fusion Transformer for Stereo Image Super-Resolution}
\begin{document}
%
\name{Hansheng Guo$^{1}$ \qquad Juncheng Li$^{2,3*}$ \qquad Guangwei Gao$^{4}$ \qquad Zhi Li$^{5}$ \qquad Tieyong Zeng$^{1*}$}
\address{$^{1}$ The Chinese University of Hong Kong, Hong Kong, China; $^{2}$Shanghai University, Shanghai, China
\\ $^{3}$Jiangsu Key Laboratory of Image and Video Understanding for Social Safety, Nanjing, China \\ $^{4}$Nanjing University of Posts and Telecommunications, Nanjing, China \\ $^{5}$East China Normal University, Shanghai, China}

\maketitle

\begin{abstract}
Stereo image super-resolution aims to boost the performance of image super-resolution by exploiting the supplementary information provided by binocular systems. Although previous methods have achieved promising results,  they did not fully utilize the information of cross-view and intra-view. To further unleash the potential of binocular images, in this letter, we propose a novel Transformer-based parallax fusion module called Parallax Fusion Transformer (PFT). PFT employs a Cross-view Fusion Transformer (CVFT) to utilize cross-view information and an Intra-view Refinement Transformer (IVRT) for intra-view feature refinement. Meanwhile, we adopted the Swin Transformer as the backbone for feature extraction and SR reconstruction to form a pure Transformer architecture called PFT-SSR. Extensive experiments and ablation studies show that PFT-SSR achieves competitive results and outperforms most SOTA methods. All code will be available.
\end{abstract}
\section{Introduction}
\label{sec:intro}

\footnotetext{Corresponding author: Juncheng Li, Tieyong Zeng}
\footnotetext{This work was supported in part by the National Key Research and Development Program of China under Project no. 2021YFE0203700, 2018AAA0100102, and 2018AAA0100100, the National Natural Science Foundation of China under Grant no. 61972212, and the YangFan Project of Shanghai under Grant no. 23YF1412800.}

Binocular cameras have been widely employed to improve the perception capabilities of vision systems in devices such as self-driving vehicles and smartphones. With the rapid development of binocular cameras, stereo image super-resolution (SSR) is becoming increasingly popular in academia and industry. Specifically, SSR attempts to reconstruct a high-resolution (HR) image from a pair of low-resolution (LR) images. With the help of additional information from a pair of binocular images at the same physical location, making full use of the information from both two images is crucial for stereo image super-resolution (SSR). 

The easiest way to implement stereo image SR is to perform single image SR (SISR) methods~\cite{dong2014learning,sisr-1-1,sisr-2-1,sisr-3-3,shi2020ddet,huang2021infrared} on stereo image pairs, respectively. These approaches, however, neglected the cross-view information between the pair of images and are incapable of reconstructing high-quality images. To address this problem, current strategies have focused on building novel cross-view feature aggregation modules, loss functions, and so on, to improve the efficiency with which image pair interaction features are used. For example, \cite{resstereo} first combined depth estimation and image resolution tasks with multiple image inputs. After that, StereoSR~\cite{wang2022ntire} took the lead in introducing CNN into Stereo SR. iPASSR \cite{ipassr} suggested a symmetric bi-directional parallax attention module (biPAM) and an inline occlusion handling scheme as its cross view interaction module to exploit symmetry cues for stereo image SR. 
Recently, several more advanced strategies for improving Stereo SR performance have been introduced. 
For instance, NAFSSR~\cite{nafssr} designed a new CNN-based backbone NAFNet~\cite{nafnet} and proposed a novel Stereo Cross Attention Module (SCAM) as parallax fusion block.
These network topologies typically included a CNN backbone for obtaining intra-view information and a parallax fusion module for combining cross-view attention. 
Since the existence of parallax, we discovered that it is also highly crucial for cross-picture features and intra-picture features to promote each other in the process of binocular feature fusion.
However, these two processes in existing works are often relatively independent, which is not conducive to the full use of image features. Meanwhile, the quality of the input features is vital for image fusion efficiency. However, existing works never consider the degree of match between the backbone networks and parallax fusion blocks. Therefore, the combination of these two pieces will be sub-optimal.

\begin{figure*}
  \centering
  \includegraphics[width=0.9\textwidth]{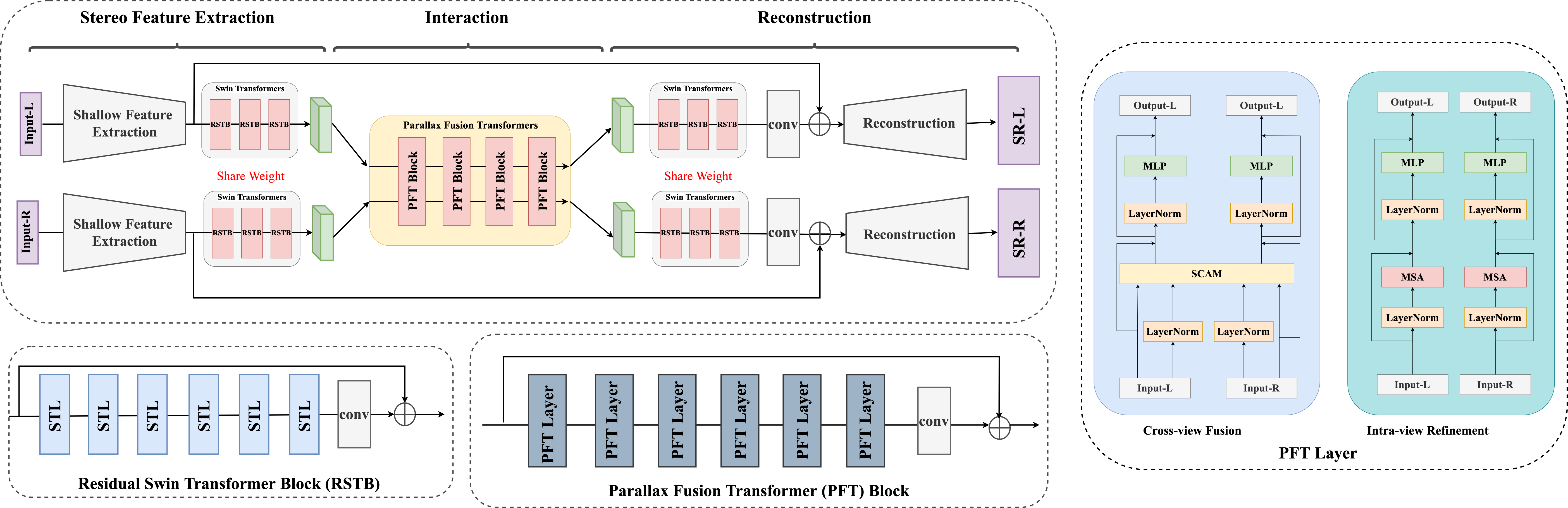} 
  \caption{The complete architecture of the proposed Parallax Fusion Transformer for Stereo Image Super-Resolution (PFT-SSR). This is a dual-stream network and interacts through an interaction module. \textbf{Due to page limit, please zoom in to see details.}}
  \label{PFT-SSR}
  \vspace{-8px}
\end{figure*}

In this work, we address the aforementioned problems by introducing the Transformer to stereo image SR. Recently, Transformer demonstrated strong performance in various low-level tasks~\cite{vit,swinir,esrt}, which can learn global information of images to further improve model performance. However, directly merging current CNN-based parallax fusion modules (PFM) and Transformer will not result in outstanding performance. This is because the CNN-based parallax fusion modules and Transformer have different properties, leading to PFM that cannot fully utilize the features from the Transformer backbone. To address this issue, we designed a new parallax fusion module, named Parallax Fusion Transformer (PFT). PFT contains a Stereo Cross Attention Module (SCAM) and a Feature Refining Module (FRM). Among them, the SCAM gets the cross-view attention and FRM will fuse the cross-view feature with the local window features. The cross-view features and intra-view features (local window features) will enhance each other to get a better representation for image super-resolution task. With the help of PFT, the proposed model can well-adapt the deep features with the parallax feature fusion blocks to fully utilize the representational potential of the Transformer. 

The contributions of this letter can be summarized as follows: 1) We propose a novel Parallax Fusion Transformer (PFT) layer with a Cross-view Fusion Transformer (CVFT) and an Intra-view Refinement Transformer (IVRT). 2) Based on the proposed PFT, we design a pure Transformer network (named PFT-SSR) to further improve the feature extraction ability of Transformer-based networks. 3) Extensive experiments have illustrated the effectiveness of PFT-SSR.

\section{METHODOLOGY}\label{Sec3}
In this paper, we propose a Parallax Fusion Transformer for Stereo Image Super-Resolution, called PFT-SSR. As shown in Fig.~\ref{PFT-SSR}, the proposed PFT-SSR consists of three parts: stereo feature extraction, feature interaction, and SR image reconstruction. For Stereo SR, the model takes two images $x_{LR}^{L}$, $x_{LR}^{R} \in R^{B \times C_{in} \times H \times W}$ as inputs and then outputs $x_{HR}^{L}$, $x_{HR}^{R} \in R^{B \times C_{out} \times S*H \times S*W}$. Among them, $B$, $C_{in}$, $C_{out}$, $H$, and $W$ are the input batch size, the number of channels, height, and weight, respectively. Meanwhile, $S$ is the upscaling factor, which is used to control the size of the output images. Specifically, we first use two convolutional layers to extract shallow features of the input images respectively. After that, we further extract the deeper feature representations with SwinIR \cite{swinir} backbone, which contains three consequent Residual Swin Transformer Blocks (RSTBs)

\begin{figure*}
 \centering
  \includegraphics[width=0.88\textwidth]{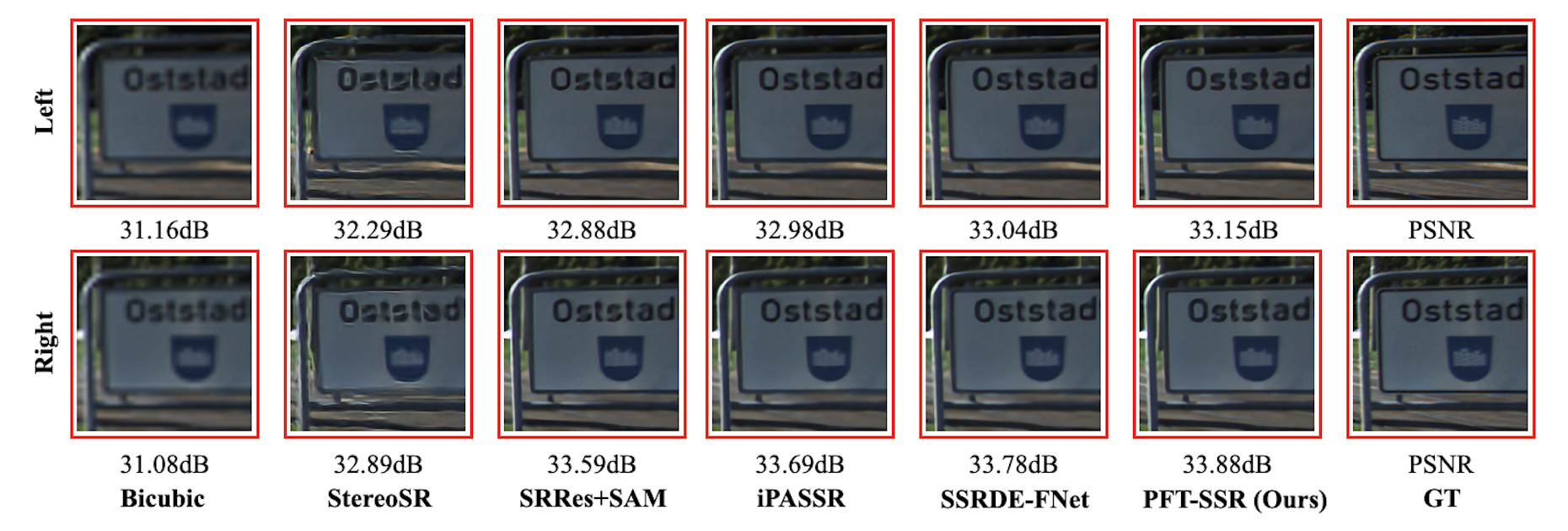} 
  \caption{Visual results (x4) achieved by different methods on Flickr1024.}
\label{Visual-1}
 \vspace{-8px}
\end{figure*}

\begin{equation}
\begin{split}
    I_d^L = f_{ex}(f_s(x_{LR}^L)), \quad I_d^R = f_{ex}(f_s(x_{LR}^R)).
\end{split}
\end{equation}
Then, the extracted features are fed into the proposed Parallax Fusion Transformers (PFT) for cross-view interaction and intra-view refinement
\begin{equation}
   I_f^L, I_f^R = f_{PFT}(I_d^R, I_d^L).
\end{equation}
With fused features, we apply RSTBs again to obtain the refined features, with a residual connection from the shallow image feature (ignored in formula for simplicity).
\begin{equation}
\begin{split}
   I_r^L= f_{cov}(f_{re}(I_f^L)), \quad
   I_r^R= f_{cov}(f_{re}(I_f^R)).
\end{split}
\end{equation}
Finally, a Reconstruction module that contains a single convolutional layer and a PixelShuffle layer is used to reconstruct the final SR images.

\begin{figure}
  \centering
  \includegraphics[width=0.3\textwidth]{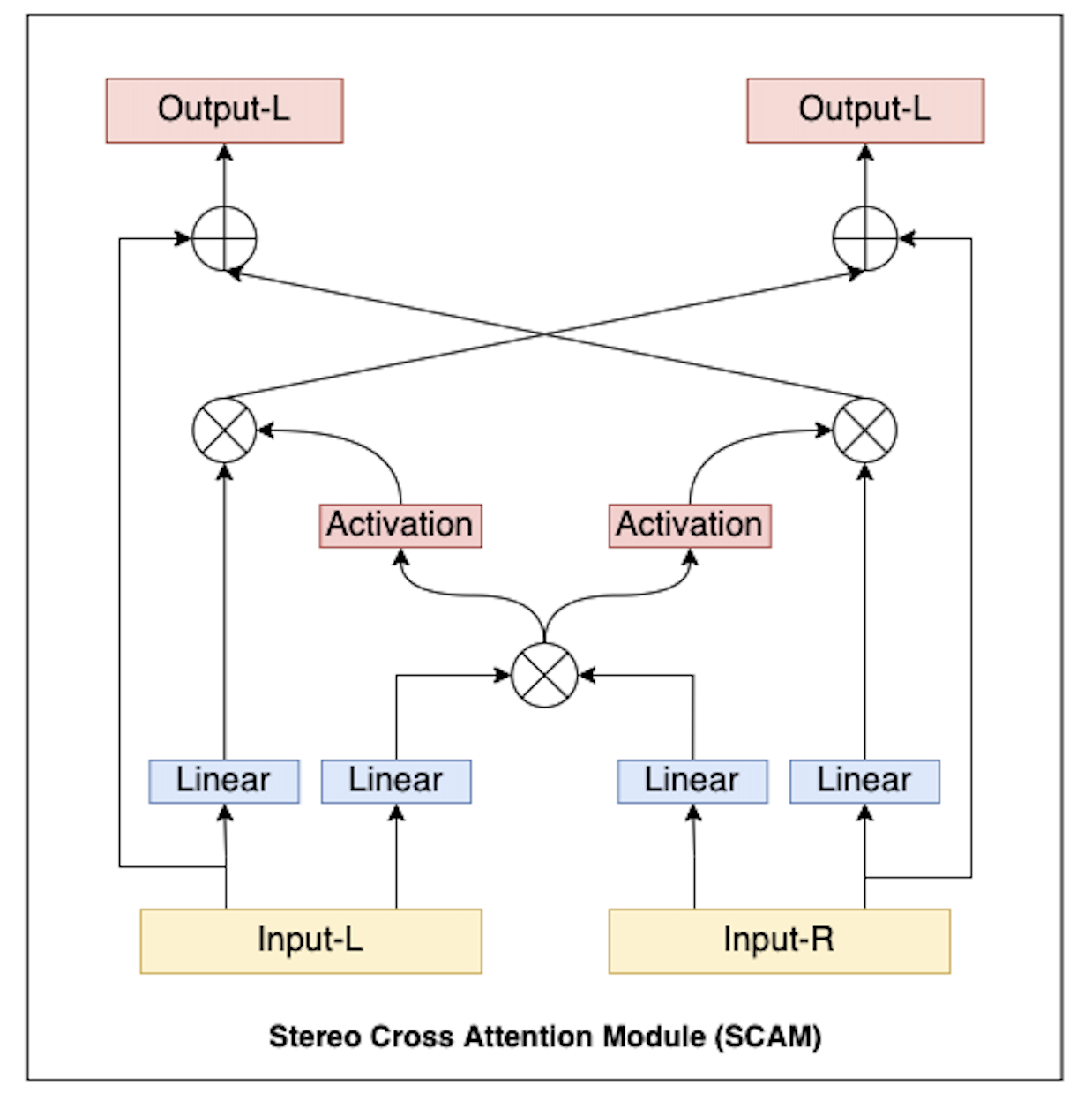} 
  \caption{The architecture of stereo cross attention module.}
  \label{SCAM}
  \vspace{-10px}
\end{figure}

\subsection{Swin Transformer Backbone}
In this work, we use Swin Transformer Blocks~\cite{swintrans} to build the backbone of our network. Specifically, a Swin Transformer Layer firstly reshapes the input feature map $I_{in}$ to $\frac{HW}{M^2}\times M^2\times C$ and performs standard self-attention locally on each window. For each of $\frac{HW}{M^2}$ feature maps, let input be $X\in {R}^{M^2\times C}$, then query, key, and value should be
\begin{equation}
    Q = XP_Q, \quad K = XP_K, \quad V = XP_V,
\end{equation}
where $P_Q$, $P_K$, and $P_V$ are linear projection matrices. Then, the attention matrix is calculated within the local windows
\begin{equation}\label{attention}
    Attention(Q, K, V) = SoftMax(QK^T/\sqrt{d}+B)V,
\end{equation}
where $B$ is the positional encoding for Transformer. The model also apply an MLP with two fully connected layers and GELU non-linearity on the attention matrix for feature transformations. Meanwhile, the LayerNorm~\cite{ba2016layer} layer is added before both Attention Block and MLP with residual connection. Though local attention can greatly reduce the amount of computation, there is no connection across local windows. To solve this problem, Swin Transformer proposed a shifted window mechanism to shifts the feature map by $(\lfloor \frac{M}{2} \rfloor, \lfloor \frac{M}{2} \rfloor)$ pixels before partitioning. The process can be expressed as 
\begin{equation}
    \begin{split}
\hspace{-0.1in}X = MSA(LN(X)) + X, X = MLP(LN(X)) + X
    \end{split}
\end{equation}
where regular partitioning and shift partitioning are used alternately before each MSA. With the help of this backbone, our model can extract sufficient useful image features.



\subsection{Parallax Fusion Transformer}
In order to make full use of the features of the left and right images, we propose a Parallax Fusion Transformer (PFT). As shown in Fig.~\ref{PFT-SSR}, PFT contains 4 PFT blocks, and each PFT block consists of 6 PFT layers and a convolutional layer. Meanwhile, each PFT layer has two different Transformer blocks, i.e. Cross-view Fusion Transformer (CVFT) and Intra-view Refinement Transformer (IVRT). Among them, CVFT adopts stereo cross-attention module (SCAM~\cite{nafssr}) to learn the features of another view and IVRT takes the local-window Transformer to better merge features from the other view to its feature map. 
Specifically, we first apply CVFT to achieve cross-view attention via SCAM. However, using single-head SCAM to get the cross-view information cannot adapt to different parallax. Therefore, we further use IVRT to make cross-view information from the other branch better interact with intra-view features. With this 'Attention-Refine' paradigm, our PFT-SSR shows a compelling effect on cross-view attention. 
\begin{table*}[t]
\centering
\scriptsize
\caption{Quantitative comparison on different datasets. PSNR$/$SSIM values achieved on both the left images (i.e., \textit{Left}) and a pair of stereo images (i.e., $\left(\textit{Left}+\textit{Right}\right)/2$) are reported. Among them, the best results are \textbf{highlighted}.} 
\label{TabQuantitative}
\renewcommand\arraystretch{1}
\resizebox{\textwidth}{!}{
\begin{tabular}{lcccccccc}
\toprule
\multirow{2}*{Method} & \multirow{2}*{Scale} & \multicolumn{3}{c}{\textit{Left}} & \multicolumn{4}{c}{$\left(\textit{Left}+\textit{Right}\right)/2$}\\
\cmidrule(lr){3-5} \cmidrule(lr){6-9}
         &           & KITTI 2012 & KITTI 2015 & Middlebury & KITTI 2012 & KITTI 2015 & Middlebury & Flickr1024 \\
\hline
EDSR~\cite{sisr-2-1} & $\times$2  & 30.83$/$0.9199 & 29.94$/$0.9231 & 34.84$/$0.9489 &30.96$/$0.9228 & 30.73$/$0.9335 & 34.95$/$0.9492 & 28.66$/$0.9087 \\
RCAN~\cite{zhang2018image} & $\times$2  & 30.88$/$0.9202 & 29.97$/$0.9231 & 34.80$/$0.9482 & 31.02$/$0.9232 & 30.77$/$0.9336 & 34.90$/$0.9486 & 28.63$/$0.9082 \\
StereoSR~\cite{StereoSR} & $\times$2  & 29.42$/$0.9040 & 28.53$/$0.9038 & 33.15$/$0.9343 & 29.51$/$0.9073 & 29.33$/$0.9168 & 33.23$/$0.9348 & 25.96$/$0.8599 \\
PASSRnet~\cite{passrnet} & $\times$2 & 30.68$/$0.9159 & 29.81$/$0.9191 & 34.13$/$0.9421 & 30.81$/$0.9190 & 30.60$/$0.9300 & 34.23$/$0.9422 & 28.38$/$0.9038 \\
iPASSR~\cite{ipassr} & $\times$2  & 30.97$/$0.9210 & 30.01$/$0.9234 & 34.41$/$0.9454 & 31.11$/$0.9240 & 30.81$/$0.9340 & 34.51$/$0.9454 & 28.60$/$0.9097 \\
SSRDE-FNet~\cite{ssrdefnet}   & $\times$2 & 31.08$/$\textbf{0.9224} & 30.10$/$\textbf{0.9245} & 35.02$/$0.9508 & 31.23$/$\textbf{0.9254} & 30.90$/$\textbf{0.9352} & 35.09$/$0.9511 & 28.85$/$\textbf{0.9132} \\
PFT-SSR (Ous)   & $\times$2  & \textbf{31.15}$/$0.9166 & \textbf{30.16}$/$0.9187 & \textbf{35.08}$/$\textbf{0.9516} & \textbf{31.29}$/$0.9195 & \textbf{30.96}$/$0.9306 & \textbf{35.21}$/$\textbf{0.9520} & \textbf{29.05}$/$0.9049 \\
\hline
EDSR~\cite{sisr-2-1} &  $\times$4  & 26.26$/$0.7954 & 25.38$/$0.7811 & 29.15$/$0.8383 & 26.35$/$0.8015 & 26.04$/$0.8039 & 29.23$/$0.8397 & 23.46$/$0.7285 \\
RCAN~\cite{zhang2018image} &  $\times$4  & 26.36$/$0.7968 & 25.53$/$0.7836 & 29.20$/$0.8381 & 26.44$/$0.8029 & 26.22$/$0.8068 & 29.30$/$0.8397 & 23.48$/$0.7286 \\
StereoSR~\cite{StereoSR}  &  $\times$4   & 24.49$/$0.7502 & 23.67$/$0.7273 &27.70$/$0.8036 & 24.53$/$0.7555 & 24.21$/$0.7511 & 27.64$/$0.8022 & 21.70$/$0.6460 \\
PASSRnet~\cite{passrnet}  &  $\times$4    & 26.26$/$0.7919 & 25.41$/$0.7772 &28.61$/$0.8232 & 26.34$/$0.7981 & 26.08$/$0.8002 & 28.72$/$0.8236 & 23.31$/$0.7195 \\
SRRes+SAM  &  $\times$4   & 26.35$/$0.7957 & 25.55$/$0.7825 & 28.76$/$0.8287 & 26.44$/$0.8018 & 26.22$/$0.8054 & 28.83$/$0.8290 & 23.27$/$0.7233 \\
iPASSR~\cite{ipassr}  &  $\times$4   & 26.47$/$0.7993 & 25.61$/$0.7850 & 29.07$/$0.8363 & 26.56$/$0.8053 & 26.32$/$0.8084 & 29.16$/$0.8367 & 23.44$/$0.7287 \\
SSRDE-FNet~\cite{ssrdefnet}  & $\times$4   & 26.61$/$\textbf{0.8028} & 25.74$/$\textbf{0.7884} & 29.29$/$0.8407 & 26.70$/$\textbf{0.8082} & 26.45$/$\textbf{0.8118} & 29.38$/$0.8411 & 23.59$/$\textbf{0.7352} \\
PFT-SSR (Ours) & $\times$4   & \textbf{26.64}$/$0.7913 & \textbf{25.76}$/$0.7775 & \textbf{29.58}$/$\textbf{0.8418} & \textbf{26.77}$/$0.7998 & \textbf{26.54}$/$0.8083 &
\textbf{29.74}$/$\textbf{0.8426} & \textbf{23.89}$/$0.7277 \\
\bottomrule
\end{tabular}}
\vspace{-8px}
\end{table*}

\textbf{Cross-view Fusion Transformer (CVFT)}:
The core component of CVFT is SCAM, and the whole process of SCAM is shown in Fig.~\ref{SCAM}. Given input image features $X_L, X_R \in {R}^{H \times W \times C}$, we first perform layer normalization to get scaled features. Due to the nature of stereo images, we use the same $Q$ and $K$ for representing intra-view features. Then, we get cross-view attention both from right to left and from left to right by 
\begin{equation}
    \begin{split}
        F_{R\rightarrow L} = Attention(T_1^L\overline{X_L},  T_1^R\overline{X_R}, T_2^R\overline{X_R}), \\ 
        F_{L\rightarrow R} = Attention(T_1^R\overline{X_R},  T_1^L\overline{X_L}, T_2^L\overline{X_L}),
    \end{split}
\end{equation}
where $Attention$ is defined same as Eq.~\eqref{attention}. Besides, $T_1^L$, $T_1^R$, $T_2^L$, and $T_2^R$ are linear projection matrices. After getting the cross-view attention feature, we use a weighted residual connection to merge it to the corresponding image feature, which are formulated as
\begin{equation}
    \begin{split}
        Y_L = \alpha_LF_{R\rightarrow L} + X_L,
        Y_R = \alpha_RF_{L\rightarrow R} + X_R,
    \end{split}
\end{equation}
where $\alpha_L$ and $\alpha_R$ are learnable scalars. After observing the corrected features, we apply MLP and LayerNorm to get the final outputs and the whole process can be expressed as
\begin{equation}
    \begin{split}
        X = SCAM(X) + X, X = MLP(LN(X)) + X.
    \end{split}
\end{equation}

\textbf{Intra-view Refinement Transformer (IVRT)}:~\label{IVRT}
One key difficulty of SSR is the different parallax brought by various stereo systems. Although SCAM shows great cross-view attention ability, it cannot adapt various parallax. After observing this, we used a Transformer with local-window attention for feature refinement. Regular partitioning is adopted before the MSA so that the features after the interaction of the two views can be further fused and enhanced, which is helpful for the final SR image reconstruction.


\section{Experiment}\label{sec4}
\subsection{Experimental Settings}
800 images from Flickr1024\cite{flickr1024} and 60 images from Middlebury\cite{middle} are chosen for training. To make the Middlebury dataset matches the spatial resolution of the Flickr1024 dataset, we perform bicubic downsampling by a factor of 2 on each image. And then, we use bicubic downsampling to these GT images by the factors of 2 and 4 to get the input images. We follow previous works~\cite{ipassr,nafssr,ssrdefnet} on this setting to make comparison fair. During training, we use the L1 loss function for supervision, PSNR and SSIM as quantitative metrics to make easy comparison with previous methods. These metrics are calculated on RGB color space with a pair of stereo images. To evaluate SR results, we use KITTI 2012~\cite{kitti2012}, KITTI 2015~\cite{kitti2015}, Middlebury~\cite{middle}, and Flickr1024~\cite{flickr1024} for test.


\subsection{Comparison to state-of-the-art methods}
We compare our proposed PFT-SSR with several state-of-the-art methods, including SISR methods (e.g., EDSR~\cite{sisr-2-1}, RCAN~\cite{zhang2018image}) and stereo image SR methods (e.g., StereoSR~\cite{StereoSR}, PASSRnet~\cite{passrnet}, iPASSR~\cite{ipassr}, and SSRDE-FNett~\cite{ssrdefnet}). According to TABLE~\ref{TabQuantitative}, we can clearly observe that our PFT-SSR achieves outstanding results and outperforms most other SOTA methods, especially on Flickr102. Meanwhile, we also show the qualitative comparisons in Figs.~\ref{Visual-1}. Obviously, our PFT-SSR can reconstruct more accurate SR images with more accurate edges and texture details. This fully demonstrates the effectiveness of the proposed PFT-SSR.

\subsection{Ablation Study}
Cross-view interaction is the key part in Stereo SR. In this part, we do ablation on the choice of this technology to show the strong stereo image fusion ability of the proposed PFT. We use Swin Transformer~\cite{swintrans} Blocks as backbones and take the same number of Swin Transformer, biPAM~\cite{ipassr}, and our proposed PFT as the cross-view interaction module in this part. According to TABLE~\ref{Ana}, it is obviously that the proposed PFT can improve the model performance more effectively, which fully illustrates the effectiveness of PFT.

\begin{table}[]
\centering
\small
\caption{Ablation study on PFT under Flickr1024.} 
\label{Ana}
\begin{tabular}{c|c|c|c}
\toprule
\multicolumn{1}{c|}{\textbf{Backbone}} & \multicolumn{1}{c|}{\textbf{Module}} & \multicolumn{1}{c|}{\textbf{PSNR (x4)}} & \multicolumn{1}{c}{\textbf{SSIM (x4)}} \\ \hline
Swin Transformer & None          & 23.54 & 0.7120 \\
Swin Transformer & RSTB (SwinIR) & 23.65 & 0.7164 \\
Swin Transformer & BiPAM         & 23.42 & 0.7068 \\
Swin Transformer & PFT (Ours)         & \textbf{23.83} & \textbf{0.7268}  \\
\bottomrule
\end{tabular}
\vspace{-8px}
\end{table}

\section{Conclusion}\label{sec5}
In this paper, we proposed a PFT-SSR for stereo image super-resolution, which contains a well-designed Parallax Fusion Transformer (PFT). PFT consists of a Cross-view Fusion Transformer (CVFT) and an Intra-view Refinement Transformer (IVRT), specially designed for cross-view interaction. It is worth mentioning that PFT can better merge different parallaxes to utilize the features of the left and right images fully. Meanwhile, PFT can also better adapt to the current popular Transformer-based backbone. Extensive experiments show that PFT-SSR outperforms most current models and achieves promising outcomes.

\bibliographystyle{IEEEbib}

\begin{thebibliography}{10}

\bibitem{dong2014learning}
Chao Dong, Chen~Change Loy, Kaiming He, and Xiaoou Tang.
\newblock Learning a deep convolutional network for image super-resolution.
\newblock In {\em ECCV}, pages 184--199, 2014.

\bibitem{sisr-1-1}
Jiwon Kim, Jung~Kwon Lee, and Kyoung~Mu Lee.
\newblock Accurate image super-resolution using very deep convolutional
  networks.
\newblock In {\em CVPR}, pages 1646--1654, 2016.

\bibitem{sisr-2-1}
Bee Lim, Sanghyun Son, Heewon Kim, Seungjun Nah, and Kyoung Mu~Lee.
\newblock Enhanced deep residual networks for single image super-resolution.
\newblock In {\em CVPR Workshops}, pages 136--144, 2017.

\bibitem{sisr-3-3}
Tao Dai, Jianrui Cai, Yongbing Zhang, Shu-Tao Xia, and Lei Zhang.
\newblock Second-order attention network for single image super-resolution.
\newblock In {\em CVPR}, pages 11065--11074, 2019.

\bibitem{shi2020ddet}
Yukai Shi, Haoyu Zhong, Zhijing Yang, Xiaojun Yang, and Liang Lin.
\newblock Ddet: Dual-path dynamic enhancement network for real-world image
  super-resolution.
\newblock {\em IEEE Signal Processing Letters}, 27:481--485, 2020.

\bibitem{huang2021infrared}
Yongsong Huang, Zetao Jiang, Rushi Lan, Shaoqin Zhang, and Kui Pi.
\newblock Infrared image super-resolution via transfer learning and psrgan.
\newblock {\em IEEE Signal Processing Letters}, 28:982--986, 2021.

\bibitem{resstereo}
Arnav~V Bhavsar and AN~Rajagopalan.
\newblock Resolution enhancement in multi-image stereo.
\newblock {\em IEEE Transactions on Pattern Analysis and Machine Intelligence},
  32(9):1721--1728, 2010.

\bibitem{wang2022ntire}
Longguang Wang, Yulan Guo, Yingqian Wang, Juncheng Li, Shuhang Gu, Radu
  Timofte, Liangyu Chen, Xiaojie Chu, Wenqing Yu, Kai Jin, et~al.
\newblock {NTIRE} 2022 challenge on stereo image super-resolution: Methods and
  results.
\newblock In {\em CVPR Workshop}, pages 906--919, 2022.

\bibitem{ipassr}
Yingqian Wang, Xinyi Ying, Longguang Wang, Jungang Yang, Wei An, and Yulan Guo.
\newblock Symmetric parallax attention for stereo image super-resolution.
\newblock In {\em CVPR Workshop}, pages 766--775, 2021.

\bibitem{nafssr}
Xiaojie Chu, Liangyu Chen, and Wenqing Yu.
\newblock Nafssr: Stereo image super-resolution using nafnet.
\newblock In {\em CVPR}, pages 1239--1248, 2022.

\bibitem{nafnet}
Liangyu Chen, Xiaojie Chu, Xiangyu Zhang, and Jian Sun.
\newblock Simple baselines for image restoration.
\newblock {\em arXiv preprint arXiv:2204.04676}, 2022.

\bibitem{vit}
Alexey Dosovitskiy, Lucas Beyer, Alexander Kolesnikov, Dirk Weissenborn,
  Xiaohua Zhai, Thomas Unterthiner, Mostafa Dehghani, Matthias Minderer, Georg
  Heigold, Sylvain Gelly, et~al.
\newblock An image is worth 16x16 words: Transformers for image recognition at
  scale.
\newblock {\em arXiv preprint arXiv:2010.11929}, 2020.

\bibitem{swinir}
Jingyun Liang, Jiezhang Cao, Guolei Sun, Kai Zhang, Luc Van~Gool, and Radu
  Timofte.
\newblock Swinir: Image restoration using swin transformer.
\newblock In {\em ICCV}, pages 1833--1844, 2021.

\bibitem{esrt}
Zhisheng Lu, Juncheng Li, Hong Liu, Chaoyan Huang, Linlin Zhang, and Tieyong
  Zeng.
\newblock Transformer for single image super-resolution.
\newblock In {\em CVPR}, pages 457--466, 2022.

\bibitem{swintrans}
Ze~Liu, Yutong Lin, Yue Cao, Han Hu, Yixuan Wei, Zheng Zhang, Stephen Lin, and
  Baining Guo.
\newblock Swin transformer: Hierarchical vision transformer using shifted
  windows.
\newblock In {\em ICCV}, pages 10012--10022, 2021.

\bibitem{ba2016layer}
Jimmy~Lei Ba, Jamie~Ryan Kiros, and Geoffrey~E Hinton.
\newblock Layer normalization.
\newblock {\em arXiv preprint arXiv:1607.06450}, 2016.

\bibitem{zhang2018image}
Yulun Zhang, Kunpeng Li, Kai Li, Lichen Wang, Bineng Zhong, and Yun Fu.
\newblock Image super-resolution using very deep residual channel attention
  networks.
\newblock In {\em ECCV}, 2018.

\bibitem{StereoSR}
Daniel~S Jeon, Seung-Hwan Baek, Inchang Choi, and Min~H Kim.
\newblock Enhancing the spatial resolution of stereo images using a parallax
  prior.
\newblock In {\em CVPR}, pages 1721--1730, 2018.

\bibitem{passrnet}
Longguang Wang, Yingqian Wang, Zhengfa Liang, Zaiping Lin, Jungang Yang, Wei
  An, and Yulan Guo.
\newblock Learning parallax attention for stereo image super-resolution.
\newblock In {\em CVPR}, pages 12250--12259, 2019.

\bibitem{ssrdefnet}
Qinyan Dai, Juncheng Li, Qiaosi Yi, Faming Fang, and Guixu Zhang.
\newblock Feedback network for mutually boosted stereo image super-resolution
  and disparity estimation.
\newblock In {\em ACMMM}, pages 1985--1993, 2021.

\bibitem{flickr1024}
Yingqian Wang, Longguang Wang, Jungang Yang, Wei An, and Yulan Guo.
\newblock Flickr1024: A large-scale dataset for stereo image super-resolution.
\newblock In {\em ICCV Workshops}, 2019.

\bibitem{middle}
Daniel Scharstein, Heiko Hirschm{\"u}ller, York Kitajima, Greg Krathwohl, Nera
  Ne{\v{s}}i{\'c}, Xi~Wang, and Porter Westling.
\newblock High-resolution stereo datasets with subpixel-accurate ground truth.
\newblock In {\em GCPR}, pages 31--42, 2014.

\bibitem{kitti2012}
Andreas Geiger, Philip Lenz, and Raquel Urtasun.
\newblock Are we ready for autonomous driving? the kitti vision benchmark
  suite.
\newblock In {\em CVPR}, pages 3354--3361, 2012.

\bibitem{kitti2015}
Moritz Menze and Andreas Geiger.
\newblock Object scene flow for autonomous vehicles.
\newblock In {\em CVPR}, pages 3061--3070, 2015.

\end{thebibliography}

\end{document}